\documentclass[letterpaper]{article} 
\usepackage{aaai25}  
\usepackage{times}  
\usepackage{helvet}  
\usepackage{courier}  
\usepackage[hyphens]{url}  
\usepackage{graphicx} 
\usepackage{natbib}  

\usepackage{caption} 
\usepackage{algorithm}
\usepackage{algorithmic}

\usepackage{inconsolata}

\usepackage{graphicx}
\usepackage{svg}

\usepackage{multirow}
\usepackage{amsmath}

\usepackage{url}            
\usepackage{booktabs}       
\usepackage{amsfonts}       
\usepackage{nicefrac}       
\usepackage{microtype}      
\usepackage{xcolor}         

\usepackage{newfloat}
\usepackage{listings}
\DeclareCaptionStyle{ruled}{labelfont=normalfont,labelsep=colon,strut=off} 
\floatstyle{ruled}
\newfloat{listing}{tb}{lst}{}
\floatname{listing}{Listing}
\title{Relation-aware Hierarchical Prompt for Open-vocabulary Scene Graph Generation}
\author{
    Tao Liu\textsuperscript{\rm 1},
    Rongjie Li\textsuperscript{\rm 1},
    Chongyu Wang\textsuperscript{\rm 1},
    Xuming He\textsuperscript{\rm 1,2}\thanks{Corresponding author}
}
\affiliations{
    \textsuperscript{\rm 1}ShanghaiTech University, Shanghai, China\\

    \textsuperscript{\rm 2}Shanghai Engineering Research Center of Intelligent Vision and Imaging\\
    liutao2023@shanghaitech.edu.cn, bugboy56@gmail.com, \{wangchy, hexm\}@shanghaitech.edu.cn
}

\usepackage{bibentry}

\begin{document}

\maketitle

\begin{abstract}
Open-vocabulary Scene Graph Generation (OV-SGG) overcomes the limitations of the closed-set assumption by aligning visual relationship representations with open-vocabulary textual representations. This enables the identification of novel visual relationships, making it applicable to real-world scenarios with diverse relationships.  However, existing OV-SGG methods are constrained by fixed text representations, limiting diversity and accuracy in image-text alignment.  To address these challenges, we propose the Relation-Aware Hierarchical Prompting (RAHP) framework, which enhances text representation by integrating subject-object and region-specific relation information.  Our approach utilizes entity clustering to address the complexity of relation triplet categories, enabling the effective integration of subject-object information. Additionally, we utilize a large language model (LLM) to generate detailed region-aware prompts, capturing fine-grained visual interactions and improving alignment between visual and textual modalities.  RAHP also introduces a dynamic selection mechanism within Vision-Language Models (VLMs), which adaptively selects relevant text prompts based on the visual content, reducing noise from irrelevant prompts.  Extensive experiments on the Visual Genome and Open Images v6 datasets demonstrate that our framework consistently achieves state-of-the-art performance, demonstrating its effectiveness in addressing the challenges of open-vocabulary scene graph generation. The code is available at: \url{https://github.com/Leon022/RAHP}
\end{abstract}

%

\section{Introduction}
\label{sec:intro}
Scene Graph Generation (SGG) \cite{johnson2015image,zellers2018neural} is a fundamental task in computer vision, involving the construction of a structured representation of a scene by identifying the relations between entities depicted in an image. It has already demonstrated promising performance in various downstream tasks \cite{kamath2021mdetr,lee2019visual,chen2020say,li2021x}. Traditional SGG methods typically operate within a closed vocabulary, and due to the diversity of relational concepts that exceed existing data annotations, they face challenges in effectively modeling open-set relations. To address this challenge, Open-Vocabulary Scene Graph Generation (OV-SGG) \cite{he2022towards,zhang2023learning,yu2023visually} has emerged as an active research area recently.
\begin{figure}[t]
  \centering
  \includegraphics[width=0.8\linewidth]{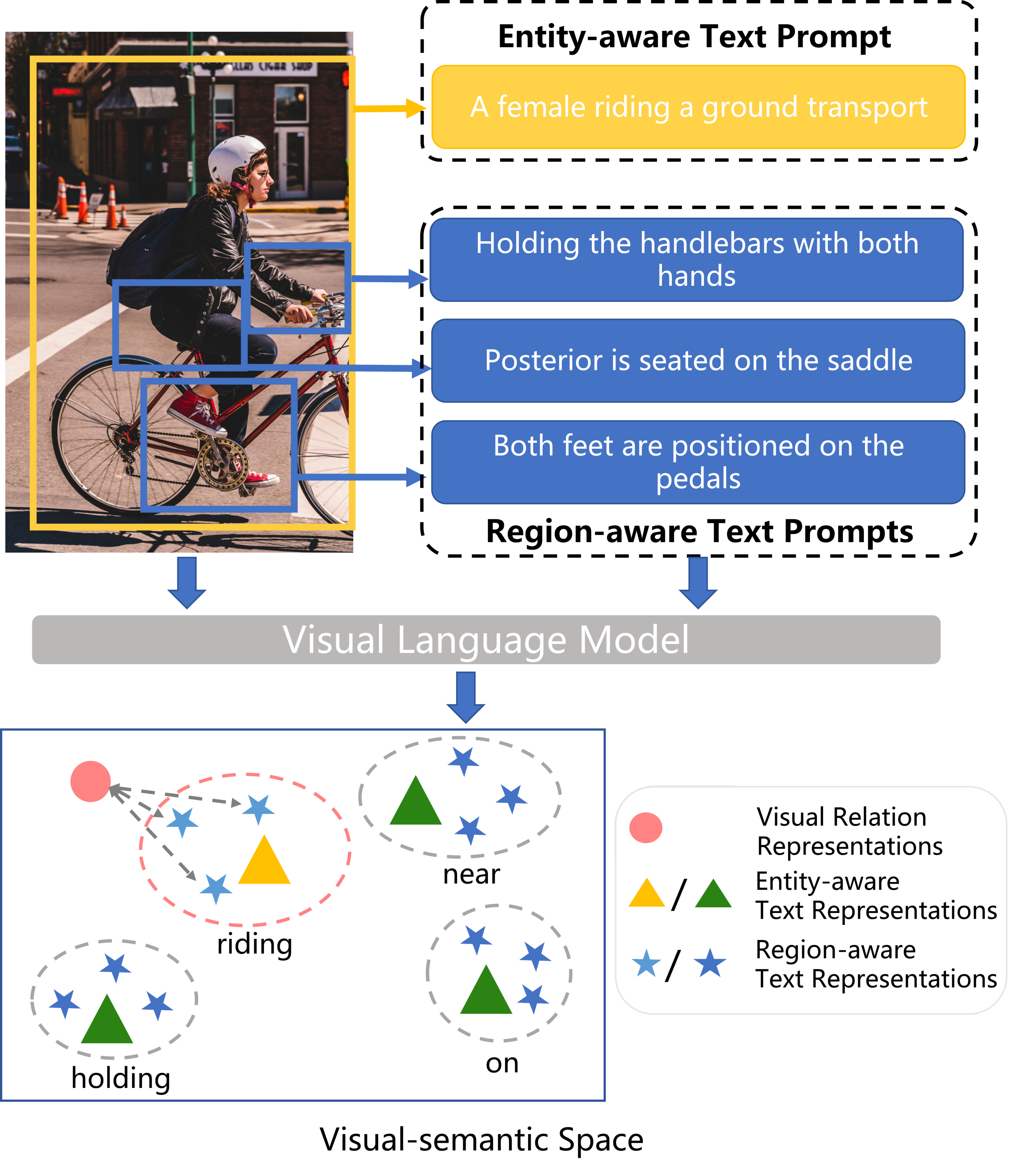}
  \caption{An illustration of RAHP for OV-SGG. RAHP generates entity-aware and region-aware hierarchical prompts to enrich the text representations of the relation, thereby enhancing OV-SGG.
  }
  \label{fig:intro_fig}
\end{figure}

Previous studies~\cite{yu2023visually,liao2022gen,chen2023expanding} mainly leverage the image-text matching capabilities of pre-trained Vision-Language Models (VLMs)~\cite{radford2021learning} to achieve open-vocabulary relation detection based on the similarity scores between relation features and text representations. However, these methods typically rely on a single, fixed form of text representation, which limits the diversity and accuracy of image-text matching, particularly in predicting novel relations. 
To address this, a promising strategy ~\cite{gao2023compositional,li2024zero,menon2022visual} is to use generative category descriptions to expand the text representation space, thereby enhancing the flexibility and precision of image-text matching. Nonetheless, generating informative relation descriptions requires not only encoding triplet information~\cite{li2024leveraging}, i.e., $<$subject, predicate, object$>$, but also capturing fine-grained interactions corresponding to different image regions. 
Given the quadratic growth in triplet combinations with increasing subjects and objects, incorporating all triplet information into representations becomes impractical.
Moreover, previous methods often employ all the generated descriptions for matching with the entities or relations in the image, which usually includes many irrelevant descriptions for the input image. This introduces a significant amount of noise in text representation, thereby reducing prediction accuracy.




To address the above challenges, we introduce the Relation-Aware Hierarchical Prompting framework (RAHP), which integrates subject-object and regional relation information within a relational representation space. 
We focus on enhancing the textual representation in the visual-semantic space of VLMs. 
As shown in Fig. \ref{fig:intro_fig}, we extend the range of relational text representations in the space, enabling the text representations to pair more effectively with visual representations. In open-vocabulary tasks, this approach can significantly enhance their consistency with visual representations and improve the model's generalization.

Specifically, RAHP contains a hierarchical prompt generation module that reduces triplet category space through entity clustering, lowering the complexity of encoding triplet information. This module also uses a large language model (LLM) to identify key regions for both subjects and objects, generating fine-grained region descriptions as text prompts by combining these regions. 
Those entity-aware and region-aware text representations more effectively captures contextual information in visual data, enhancing the model's understanding and generalization in complex interactive scenes.
Additionally, RAHP implements a VLM-based dynamic selection mechanism that filters out completely irrelevant text representations based on visual concept, thereby improving matching accuracy.


We conduct extensive experiments to validate our approach on two SGG benchmarks: Visual Genome~\cite{krishna2017visual} and Open Images-v6~\cite{kuznetsova2020open} datasets, and it achieves state-of-the-art performance.

The main contribution of our work has three folds.

\begin{itemize}
\item We propose a relation-aware hierarchical prompting framework (RAHP) for OV-SGG that integrates entity-aware and region-aware text prompts, enhancing text representations and model generalization.
\item We introduce a VLM-guided dynamic selection mechanism that adapts text prompts based on visual information, minimizing irrelevant content and enhancing the robustness of relation predictions.
\item Experiments on two benchmark datasets demonstrate that our method achieves state-of-the-art generalization performance in OV-SGG.
\end{itemize}

\section{Related Work}
\label{sec:related}

\subsection{Scene Graph Generation}
\label{sec:related-1}

The Scene Graph Generation (SGG) task, initially proposed by \cite{johnson2015image}, traditionally relies on supervised methods~\cite{xu2017scene,gu2019scene,tang2019learning,zellers2018neural} that predict relationships using visual, spatial, and contextual cues. 
To reduce reliance on annotated scene graphs, some approaches \cite{suhail2021energy,yang2019auto} use language supervision, extracting entity and relationship labels from image captions \cite{li2022integrating,shi2021simple,zhonglearning}. However, these methods often use closed-set classifiers, limiting their ability to handle novel entities or relations.
Recent studies have expanded SGG to open-vocabulary settings \cite{he2022towards,zhang2023learning,yu2023visually}. For example, SVPR \cite{he2022towards} uses dense caption pre-training and prompt fine-tuning, while $\mathrm{VS^3}$ \cite{zhang2023learning} aligns visual features with a pre-trained visual-semantic space for predicting new entities. Epic~\cite{yu2023visually} introduces cross-modal entanglement, combining text and region embeddings to classify new predicates. OvSGTR~\cite{chen2023expanding} extends OV-SGG to open-vocabulary detection and relations-based scenarios.

Most of these methods rely on visual-text matching to classify novel relations, using fixed-form text prompts that limit recognition of novel relations in OV-SGG. To address this, we propose a hierarchical text representation enhancement method that enriches the text representation space by introducing text prompts at both subject-object and region levels, improving relationship recognition.

\subsection{Open-vocabulary Methods}

In recent years, researchers in the field of visual scene understanding, such as object detection \cite{gu2021open, zareian2021open}, have shifted their focus from traditional closed-set methods to more flexible open-vocabulary methods.
A key driver of this evolution is the development and maturity of VLMs \cite{radford2021learning, jia2021scaling, li2023blip}. These models are typically pre-trained on large-scale image-text pairs, endowing them with strong cross-modal alignment capabilities. By leveraging natural language prompts \cite{wu2024towards}, VLMs can compute similarities between images and language in open-vocabulary settings, facilitating category expansion \cite{gu2021open, ma2022open}.

Early research \cite{wang2023hierarchical, menon2022visual} concentrates on simple prompts for open-vocabulary recognition. Other approaches \cite{wang2024learning, wang2022hpt} employ learnable prompts to enhance image classification. As research progresses, single prompts cannot adequately handle complex visual inputs, leading to the proposal of hierarchical prompting methods to better structure intricate query information. Models like \cite{ge2023improving} rely on object class hierarchies by WordNet \cite{miller1995wordnet}. RECODE \cite{li2024zero} utilizes LLMs to generate hierarchical prompts from the perspectives of subject, object, and spatial levels, facilitating zero-shot relationship recognition. 
In contrast to RECODE, our work approaches the task from a regional perspective, enabling the generation of more detailed and specific relationship prompts.
Additionally, we introduce a dynamic VLM-guided mechanism that adjusts prompts based on visual inputs, increasing the accuracy and flexibility of text representations.


\section{Preliminary}
\label{sec:preliminary}

\subsection{Problem Setting}
\label{sec:preliminary-1}
 The goal of SGG is to create a descriptive graph $\mathcal{G}=\left \{\mathcal{V}, \mathcal{R} \right \}$ from an image $I$. This graph consists of $N^v$ entities $\mathcal{V}=\{\mathbf{v}_i\}_{i=1}^{N^v}$ and visual relationship triples $\mathcal{E}=\{\mathbf{v}_i,r_{i,j},\mathbf{v}_j\}_{i\neq j}$, where $r_{i,j}$ represents the predicate category between them. Each entity $\mathbf{v}_i$ is represented as $(c^v_i,\mathbf{b}_i)$, where $c^v_i$ denotes the label in the entity category space $\mathcal{O}^e$, and $\mathbf{b}_i$ represents its location through a bounding box in the image. The predicate category $r_{i,j}$ denotes the label in the category space $\mathcal{O}^r$. In the task of OV-SGG, the category spaces for entities and predicates are divided into two parts. Specifically, the predicate category space contains the base category space $\mathcal{O}_b^r$ and the novel category space $\mathcal{O}_n^r$, and it has $\mathcal{O}^r = \mathcal{O}_b^r \cup  \mathcal{O}_n^r$. Similarly, the entity category space also has $\mathcal{O}^e = \mathcal{O}_b^e \cup  \mathcal{O}_n^e$.

\subsection{OV-SGG Pipeline}
\label{sec:preliminary-2}

Most OV-SGG methods~\cite{yu2023visually,chen2023expanding} can typically be decoupled into two steps: relationship proposal generation and predicate classification. 

First, the model receives an image as input and feeds it into a proposal network, from which it extracts relationship proposals $\mathcal{P}=\{\mathbf{v}_i,\mathbf{v}_j\}_{i\neq j}$ and the corresponding relationship features $\mathbf{R} \in \mathbb{R}^{N \times d}$, where $N$ is the number of relationship proposals and $d$ is the dimension of the feature representation.

Then, the relationship features are fed into the predicate classifier as visual representations. The predicate classifier usually handles each predicate class using predefined text prompts, which generate text embeddings $\mathbf{T} \in \mathbb{R}^{\mathbb{C}_p \times d}$  through the text encoder $\mathrm{TextEnc}$ of a VLM, where $\mathbb{C}_p$ is the number of predicate categories. These text embeddings as the text representations replace the fixed predicate classifier weights, enabling the model to extend to new relationship categories that appear during the testing phase.
The predicate classifier obtains the predicate classification scores $\mathbf{S} \in \mathbb{R}^{N \times \mathbb{C}_p}$ for each relationship proposal by calculating the similarity score between $\mathbf{R}$ and $\mathbf{T}$:
\begin{equation}
\mathbf{S} = \phi(\mathbf{R},\mathbf{T}) = \frac{\mathbf{R}\cdot \mathbf{T}}{\left | \mathbf{R} \right | \cdot \left | \mathbf{T} \right | },
\label{equ: predicate score}
\end{equation}
where · is the dot product, we define this operation of calculating similarity as $\phi ()$.
During OV-SGG training, OV-SGG methods use a distillation loss to distill the knowledge of the VLM to maintain the model's generalization. The distillation loss ensures that the distance between the text embeddings and relationship features remains consistent across all pairwise classifications.

\section{Method}
\label{sec:method}
\subsection{Method Overview}

We propose RAHP, a method that enhances the generalization of OV-SGG models on novel relations by using multi-level text prompts to strengthen visual relation text representations.
Specifically, our framework is composed of three modules: hierarchical prompt generation (Sec.~\ref{sec:method-1}), visual relationship extraction (Sec.~\ref{sec:method-2}), and hierarchical relationship prediction  (Sec.~\ref{sec:method-3}).
Finally, we introduce the learning and inference pipeline of our method (Sec.~\ref{sec:method-4}).


\subsection{Hierarchical Prompt Generation}
\label{sec:method-1}

\begin{figure*}[tb]
  \centering
  \includegraphics[width=\textwidth]{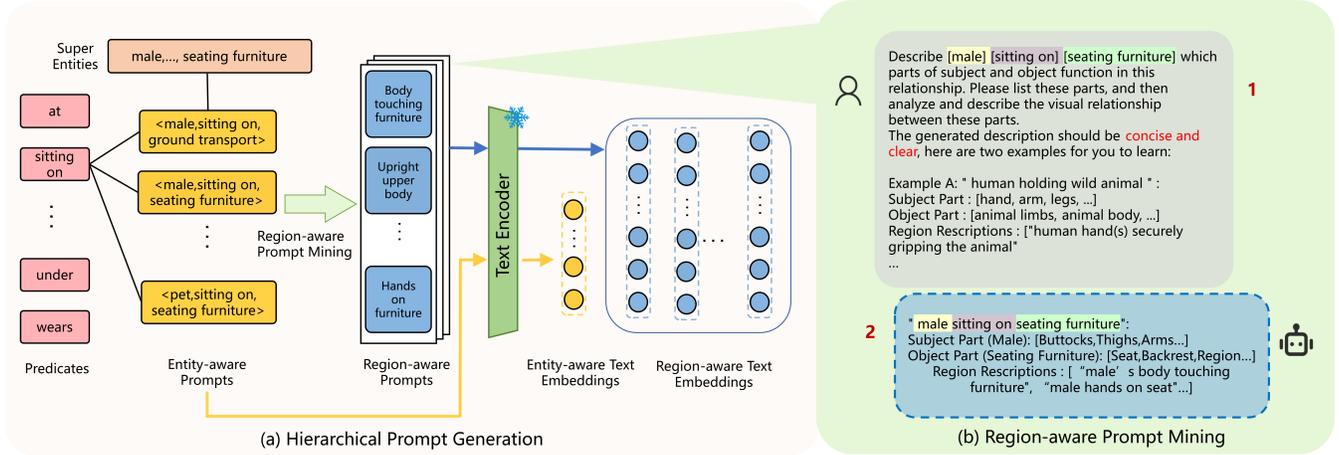}
  \caption{An overview of hierarchical prompt generation: (a) Predicates combine with super entities to create entity-aware prompts, which then expand into region-aware prompts. This process builds a rich textual representation space for extended relational triplets. (b) In region-aware prompt mining, our approach guides an LLM to decompose subjects and objects into distinct parts, enabling more detailed regional visual relationship descriptions.
  }
  \label{fig:constrcution_fig}
\end{figure*}

The hierarchical prompt generation module enriches text representations by creating multi-level text prompts that include entity-aware and region-aware prompts. 
As shown in Fig.~\ref{fig:constrcution_fig} (a), for the input vocabularies, we sequentially generate prompts at two levels.

\begin{itemize}
\item \textbf{Entity-aware text prompts}: These prompts include precise relationship content by combining predicate, subject, and object details. However, as the number of triplets grows cubically with subjects and objects, incorporating all triplet information into the prompts becomes impractical. To address this, we first cluster entities into super entities based on similarity. Similar to the approach in~\cite{zhang2024hiker}, it can effectively reduce the triplet category space (more details can be found in the appendix). We then generate entity-aware text prompts by combining super entities with predicate categories.

\item \textbf{Region-aware text prompts}: Building on entity-aware prompts, we create region-aware text prompts that capture finer visual details through a region-aware description mining strategy. As shown in Fig.~\ref{fig:constrcution_fig} (b), we use an LLM to decompose key entity parts and naturally generate region-level visual relation descriptions by combining these parts' relationships. \textit{For example, in the relationship triplet $<$male, sitting on seating, furniture$>$ the ``male'' can be associated with specific body parts like the hips, thighs, and arms, while the ``seating furniture'' can be associated with components like the seat and backrest. The ``sitting on'' relationship is then represented by combining these elements in the LLM to provide extensive region-aware relation descriptions.} Following ~\cite{menon2022visual}, we design two cases for the LLM to learn from.
\end{itemize}

\subsubsection{Hierarchical Prompt Encoding} After generating the two levels of prompts, we encode them into text embeddings as text representations using the frozen VLM text encoder $\mathrm{TextEnc}$. As shown in Fig.~\ref{fig:constrcution_fig} (a), we generate sentences for the entity-aware prompts through the template \textit{``A photo of a/an [Subject] [Predicate] a/an [Object]''}. The entity-aware prompts are encoded by $\mathrm{TextEnc}$ into an entity-aware text embedding set $\mathcal{T}^e = \{ \mathbf{T}^e_1,\mathbf{T}^e_2,...,\mathbf{T}^e_{\mathbb{C}_{se}^2} \}$, where $ \mathbf{T}^e  \in \mathbb{R}^{\mathbb{C}_p \times d}$, $\mathbb{C}_p$ represents the number of predicate categories, and $\mathbb{C}_{se}$ denotes the number of super entity categories. Correspondingly, the region-aware prompts are generated sentences through the template \textit{``A region that reflects [region descriptions]''}. The region-aware prompts are encoded into an text embedding set $\mathcal{T}^r=\{\mathbf{T}^r_1,\mathbf{T}^r_2,...,\mathbf{T}^r_{\mathbb{C}_{se}^2}\}$, where $\mathbf{T}^r_j \in \mathbb{R}^{\mathbb{C}_p \times N^r_j \times d}$, $N^r_j$ is the number of region-aware prompts, varying with the region descriptions per triplet.

\begin{figure*}[tb]
  \centering
  \includegraphics[width=0.75\textwidth]{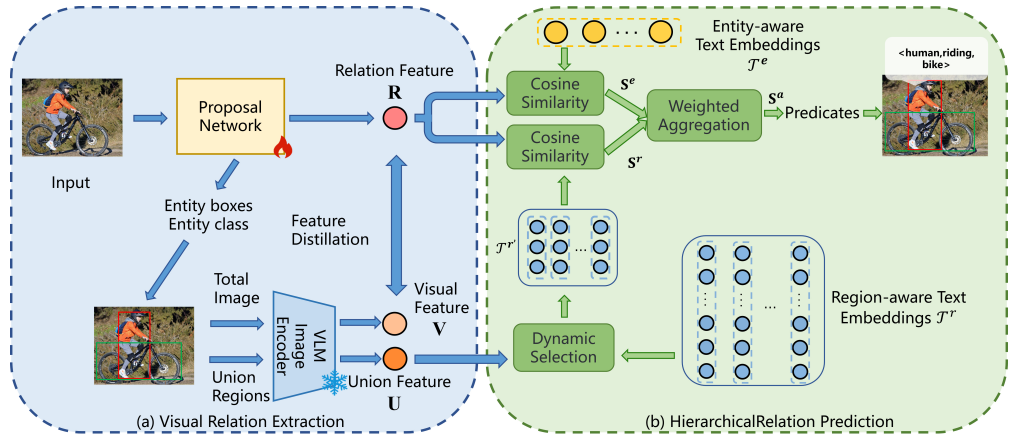}
  \caption{An overview of RAHP. 
  (a) Visual Relation Extraction Module: The process begins with extracting relation proposals and their features from the image, which are then encoded into visual features using a VLM. (b) Hierarchical Relation Prediction Module: The visual features undergo a guided selection process, where the selected embeddings are combined with entity-aware embeddings to predict predicates.
  }
  \label{fig:method_fig}
\end{figure*}

\subsection{Visual Relation Extraction}
\label{sec:method-2}
Following the process described in Sec.~\ref{sec:preliminary-2}, the visual relationship extraction module is mainly designed to extract visual relation features.
It employs a proposal network to extract relation proposals $\mathcal{P}=\{\mathbf{v}_i,\mathbf{v}_j\}_{i\neq j}$ from the visual input $I$, along with their corresponding relation feature representations $\mathbf{R}$.
Then we merge the predicted subject and object boxes in $\mathcal{P}$ into union boxes, and crop the corresponding region $I^u$ from the image $I$. $I^u$ is encoded into a unified feature $\mathbf{U} \in \mathbb{R}^{N \times d}$ by the VLM’s visual encoder $\mathrm{VisEnc}$, calculated as follows:
\begin{equation}
\mathbf{U} = \mathrm{VisEnc}(I^u).
\label{equ: VLM visual encode}
\end{equation}
Both the relation features $\mathbf{R}$ and the union features $\mathbf{U}$ are input into the hierarchical relation prediction module.

\subsection{Hierarchical Relation Prediction}
\label{sec:method-3}
The hierarchical relation prediction module predicts predicates by calculating the similarity between relation features $\mathbf{R}$ and two levels of text embeddings $\mathbf{T}$: entity-aware and region-aware. This module includes two key components: VLM-guided dynamic selection, which filters out irrelevant prompts, and hierarchical prediction aggregation, ensuring accurate predicate classification.

\subsubsection{Image-guide Dynamic Selection}  
The VLM-guided dynamic selection mechanism utilizes the image-text alignment capabilities of a VLM to match $\mathcal{T}^r$ with union features $\mathbf{U}$.
The mechanism is aimed at filtering out region-text pairs that are completely irrelevant to the image, leveraging the robust object recognition capabilities of the VLM to achieve this goal.
Specifically, for $j^{th}$ in $\mathcal{T}^r$, upon receiving the unified feature $\mathbf{U}$, it computes the matching score $\mathbf{S}^{se}_{j} \in \mathbb{R}^{N \times N^r_j}$ between $\mathbf{U}$ and the region-aware text embeddings $\mathbf{T}^r_j$ as follows:
\begin{equation}
\mathbf{S}^{se}_{j} = \phi (\mathbf{U}, \mathbf{T}^{r}_{j}) 
\label{equ: VLM prompt selection}
\end{equation}
To capture core visual semantic information, we select the top k region-aware text embeddings with the highest matching scores and perform predicate classification.
After performing VLM-guided selection on all region-aware prompts, we obtain the final region-aware text prompt embedding set ${\mathcal{T}^r}^{\prime}=\{{\mathbf{T}^r_1}^{\prime},{\mathbf{T}^r_2}^{\prime},...,{{\mathbf{T}^r}^{\prime}_{\mathbb{C}_{se}^2}}\}$, where ${\mathbf{T}^r}^{\prime} \in \mathbb{R}^{\mathbb{C}_p \times k \times d}$.
This mechanism dynamically selects text prompts based on union features, prioritizing region-aware prompts with higher probabilities for subsequent predicate prediction, effectively reducing noise.

\subsubsection{Hierarchical Prediction Aggregation}
After selecting region-aware text embeddings, we predict the final predicate scores by integrating entity-aware and region-aware embeddings. First, we calculate the similarity between $\mathbf{T}^e$ and $\mathbf{R}$ to derive the entity-aware predicate score $\mathcal{S}^e = \{ \mathbf{S}^e_1,\mathbf{S}^e_2,...,\mathbf{S}^e_{\mathbb{C}_{se}^2} \}$ for ${\mathcal{T}^e}$:
\begin{equation}
\mathbf{S}^{e}_j = \phi (\mathbf{R}, \mathbf{T}^e_j),
\label{equ: entity-aware computation}
\end{equation}
where $\mathbf{S}^e_j \in \mathbb{R}^{N \times \mathbb{C}_p}$. This score encapsulates the relation details between specific entity pairs. 
Next, at the region-aware level, we compute the region-aware predicate scores $\mathcal{S}^r = \{ \mathbf{S}^r_1,\mathbf{S}^r_2,...,\mathbf{S}^r_{\mathbb{C}_{se}^2} \}$ by evaluating the similarity between ${\mathbf{T}^r_j}^{\prime}$ and $\mathbf{R}$:
\begin{equation}
\mathbf{S}^{r}_j = \frac{\sum_{m=1}^{k} \phi (\mathbf{R}, {{\mathbf{T}^r}^{\prime}_{j,m}})}{k},
\label{equ: region-aware computation}
\end{equation}
where $\mathbf{S}^r_j \in \mathbb{R}^{N \times \mathbb{C}_p}$. The score emphasizes region relation features, providing additional text information to assist the model in understanding visual relationships. We then combine $\mathbf{S}^e_j$ and $\mathbf{S}^r_j$ to $\mathbf{S}^a_j \in \mathbb{R}^{N \times \mathbb{C}_p}$ using a weighted sum to produce the aggregated score:
\begin{equation}
\mathbf{S}^a_j = (1-\alpha) \times \mathbf{S}^{e}_j + \alpha \times \mathbf{S}^{r}_j.
\label{equ: final score computation}
\end{equation}
Finally, we select the highest scores from the $\mathbb{C}_{se}^2$ aggregated scores $\mathbf{S}^a$ as the final predicate prediction scores $\mathbf{S} \in \mathbb{R}^{N \times \mathbb{C}_p}$:
\begin{equation}
\mathbf{S} = \max (\mathbf{S}^a_{1}, \mathbf{S}^a_{2},...., \mathbf{S}^a_{\mathbb{C}_{se}^2}).
\label{equ: max final score}
\end{equation}
The prediction scores allow us to derive the probability for each predicate, enabling the determination of the predicate category.
This multi-level prediction mechanism enhances RAHP by learning regional-level text representations, improving open-vocabulary capabilities, and enabling knowledge transfer to new relationship concepts.

\subsection{SGG Learning and Inference}
\label{sec:method-4}

\subsubsection{SGG Learning}
During the training stage, the model only receives information from the base classes.
Similar to \cite{li2024sgtr,chen2023expanding}, we adopt a multi-task loss for our model training. Specifically,
we use L1 loss and GIOU loss for entity bounding box regression to reduce the gap between the predicted bounding box $\mathbf{b}$ and the ground truth $\mathbf{b}_{gt}$:
\begin{equation}
\mathcal{L}_{bbox}= \left \| \mathbf{b} - \mathbf{b}_{gt} \right \|_1 + \mathrm{GIOU}(\mathbf{b}, \mathbf{b}_{gt}).
\label{equ: bbox loss}
\end{equation}
We also use a cross-entropy loss $\mathcal{L}_{ent}=\mathrm{CE}(c^v, c^v_{gt})$ to ensure the accuracy of the prediction $c^v$ for entity classification against the ground truth category $c^v_{gt}$.

For predicate prediction, we use $\mathcal{L}_{pre}=\mathrm{FL}(r, r_{gt})$ to represent the Focal loss for predicate categories, where $r_{gt}$ is the ground truth predicate category, and $r$ is the predicted predicate category. In addition, we employ an L1 loss~\cite{liao2022gen,chen2023expanding} to minimize the gap between the relation feature $\mathbf{R}$ and the visual features $\mathbf{V} \in \mathbb{R}^{d}$ extracted by the VLM visual encoder $\mathrm{VisEnc}$. The goal is to align the relation features extracted by SGG with the VLM space, thereby enabling the prediction of novel predicates.
It also acts as a form of regularization to prevent overfitting to the specific training data. 
For the $i$-th relation proposal, the distillation loss is designed as an L1 distance loss, defined as follows:
\begin{equation}
\mathcal{L}^d_i =  \left \| {\mathbf{R}_i - \mathbf{V}} \right\|_1.
\label{equ: distillation loss}
\end{equation}

The total training loss can be written as
\begin{equation}
\mathcal{L} = \mathcal{L}_{bbox} + \lambda_1 \mathcal{L}_{ent} + \lambda_2 \mathcal{L}_{pre} + \lambda_3 \mathcal{L}^d.
\label{equ: total loss}
\end{equation}
where the weights of each loss term $\lambda_1,\lambda_2,\lambda_3$  balance the learning progress and importance across different tasks.

\subsubsection{SGG Inference}
To enhance the interpretability of novel relation triplets, we employ LLMs to generate informative visual descriptions before the inference phase.
In the post-processing stage, we systematically eliminate invalid self-connected edges and exclude triplets where subject and object entities are identical. Subsequently, the remaining triplets are ranked based on the combined scores from entity predictions and predicate predictions. The top $M$ relation triplets are then selected as the final output, providing comprehensive information in terms of subject entity probabilities, object entity probabilities, and predicate probabilities.

\begin{table*}[ht]
\centering
\setlength{\tabcolsep}{1mm}
\fontsize{9pt}{9pt}\selectfont
\begin{tabular}{l|l|l|l|l|cc|cc|cc}
\toprule
                          &                             &                            &                                &                               & \multicolumn{2}{c|}{Total (Relation)}       & \multicolumn{2}{c|}{Base (Relation)}        & \multicolumn{2}{c}{Novel (Relation)}        \\
\multirow{-2}{*}{S} & \multirow{-2}{*}{T}      & \multirow{-2}{*}{B} & \multirow{-2}{*}{D}     & \multirow{-2}{*}{M}       & R@50/100             & mR@50/100            & R@50/100             & mR@50/100            & R@50/100             & mR@50/100            \\ \midrule
                          &                             & ViT                        & DETR*                          & PGSG                         & 26.90/33.90          & 10.80/13.90          & -                    & -                    & -                    & 5.20/7.70            \\ \cline{3-4}
                          &                             &                            &                                &  SGTR$^\dagger$  & 36.32/41.51          & 13.30/17.60          & 40.50/46.67          & 19.57/24.96          & 0.00/0.00            & 0.00/0.00            \\
                          &                             &                            &                                & SGTR$^\dagger$+p    & 39.48/45.27          & 15.93/21.09          & 40.78/46.82          & 19.88/24.32          & 10.81/18.41          & 9.00/13.15           \\
                          & \multirow{-4}{*}{o} & \multirow{-3}{*}{R-101}    & \multirow{-3}{*}{DETR}         &  \textbf{SGTR$^\dagger$+RAHP} & \textbf{39.92/46.03} & \textbf{16.88/22.18} & \textbf{41.29/47.65} & \textbf{20.51/25.18} & \textbf{15.46/20.37} & \textbf{11.82/15.46} \\ \cmidrule{2-11} 
                          &                             & R-50                       &                                & SVPR                         & 33.50/35.90          & 8.30/10.80           & -                    & -                    & -                    & -                    \\ \cline{3-3}
                          &                             &                            &                                & Epic                          & -                    & 16.50/21.80          & 28.30/31.10          & -                    & 13.90/18.30          & -                    \\
                          &                             &                            &                                & PE-NET                        & 58.79/61.23          & 19.18/20.97          & 63.62/67.09          & 23.18/25.79          & 0.00/0.00            & 0.00/0.00            \\
                          &                             &                            &                                & PE-NET+p                           & 62.21/67.25          & 21.94/27.91          & 62.73/67.76          & 22.23/28.02          & 17.62/25.67          & 12.93/19.32          \\
\multirow{-9}{*}{PredCLS} & \multirow{-5}{*}{t} & \multirow{-4}{*}{R-101}    & \multirow{-5}{*}{ \shortstack{Faster \\ R-CNN}} & \textbf{PE-NET+RAHP}                        & \textbf{64.70/69.11} & \textbf{24.50/28.25} & \textbf{65.15/70.54} & \textbf{24.99/30.19} & \textbf{20.79/29.00} & \textbf{15.70/23.73} \\ \midrule
                          &                             &                            &                                & OvSGTR                        & 20.46/23.86          & 3.91/4.62                    & 26.14/\textbf{30.16}                    & 4.81/5.60                    & 13.45/16.19          & 1.82/2.32                    \\
\multirow{-2}{*}{SGDet}   & \multirow{-2}{*}{o} & \multirow{-2}{*}{Swin-T}   & \multirow{-2}{*}{DETR}         & \textbf{OvSGTR+RAHP}                         & \textbf{21.50/25.74} & \textbf{4.51/5.37}            & \textbf{26.29/30.16}                    & \textbf{5.15/5.94}                    & \textbf{15.59/19.92} & \textbf{3.01/4.04}                    \\ \bottomrule
\end{tabular}%
\caption{Experimental results of OVR-SGG on VG test set. - to signify methods that did not produce the result, p indicates the use of fixed-format text prompts, while DETR* denotes models with structural modifications.
S is the SGG setting; T denotes the SGG model type, o means one-stage model, t is two-stage model; B is the backbone model; D is the object detector; M represents the model.
}
\label{tab:main-table}
\end{table*}

\begin{table*}[ht]
\centering
\setlength{\tabcolsep}{1mm}
\fontsize{9pt}{9pt}\selectfont
\begin{tabular}{l|l|l|l|l|cc|cc|cc}
\toprule
\multirow{2}{*}{S} & \multirow{2}{*}{T}      & \multirow{2}{*}{B} & \multirow{2}{*}{D} & \multirow{2}{*}{M} & \multicolumn{2}{c|}{Total (Relation)}       & \multicolumn{2}{c|}{Base (Relation)}        & \multicolumn{2}{c}{Novel (Relation)}        \\
                         &                            &                           &                           &                        & R@50/100             & mR@50/100            & R@50/100             & mR@50/100            & R@50/100             & mR@50/100            \\ \midrule
\multirow{4}{*}{SGDet}   & \multirow{4}{*}{o} & ViT                       & DETR*                     & PGSG                   & 41.30/43.30          & 20.80/23.00          & -                    & -                    & -                    & 3.80/8.90            \\ \cline{3-4}
                         &                            & \multirow{3}{*}{R-101}    & \multirow{3}{*}{DETR}     & SGTR$^\dagger$                  & 36.10/38.40          & 11.00/16.70          & -                    & -                    & -                    & 0.00/0.00            \\
                         &                            &                           &                           & SGTR$^\dagger$+p                    & 60.48/62.63          & 29.09/31.44          & 74.11/77.17          & 36.32/37.43          & 46.31/53.76          & 27.82/32.66          \\
                         &                            &                           &                           & \textbf{SGTR$^\dagger$+RAHP}                  & \textbf{62.42/64.86} & \textbf{30.79/34.46} & \textbf{78.21/80.27} & \textbf{37.72/38.66} & \textbf{49.61/56.28} & \textbf{29.43/34.16} \\ \bottomrule
\end{tabular}%
\caption{Experimental results of OVR-SGG on OIV6 test set. - to signify methods that did not produce the result, p indicates the use of fixed-format text prompts, while DETR* denotes models with structural modifications. 
S is the SGG setting; T denotes the SGG model type, o means one-stage model; B is the backbone model; D is the object detector; M represents the model.}
\label{tab:OIV6-table}
\end{table*}

\section{Experiment}
\label{sec:experiment}
In this section, we comprehensively evaluate our RAHP on the OV-SGG task. More results, including closed-set SGG, parameter sensitivity experiments and qualitative analysis, are provided in the Appendix.

\subsection{Datasets and Experimental Settings}
\label{sec:dataset}

\subsubsection{Datasets}
To evaluate the SGG task, we adopt two benchmarks: the VG150 version of the Visual Genome (VG) dataset \cite{krishna2017visual} and the Open Image v6 (OIV6) dataset \cite{kuznetsova2020open}.

\subsubsection{Evaluation metrics}
We evaluate our method under two settings~\cite{chen2023expanding}:  Open Vocabulary Relation-based Scene Graph Generation (OVR-SGG) uses a closed vocabulary for objects and an open one for relationships, whereas Open Vocabulary Detection + Relation-based Scene Graph Generation (OVD+R-SGG) uses open vocabularies for both. We adopt the PredCLS and SGDet protocols~\cite{xu2017scene} and report the performance on Recall @K (K=50/100) and mean Recall @mK (mK=50/100) for each setting.

\subsubsection{Implementation Details}
We employ the GPT-3.5-turbo, as our LLM.  
We adopt CLIP \cite{radford2021learning} (ViT-B/32) as our VLM backbone.
We categorize 150 entities into 30 super-class entities for VG and categorized 602 entities into 53 super-class entities for OIV6 (details can be found in the appendix).
RAHP is applicable to both one-stage and two-stage models, therefore we select the one-stage methods SGTR$^\dagger$~\cite{li2024sgtr} and OvSGTR (Swin-T)~\cite{chen2023expanding}, as well as the two-stage methods PR-NET~\cite{zheng2023prototype} and $\mathrm{VS^3}$ (Swin-T)~\cite{zhang2023learning} to validate the generality and high adaptability of RAHP. In each method, we retain the visual module and proposal network, replace the predicate classification part with RAHP's OV predicate prediction approach.
To align relation and VLM features, we equip CLIP with a three-layer MLP of size 512. We set $k$ = 3 to dynamically select and set $\alpha$ = 0.25 to balance the weights of the two text prompts.
For training losses, the weight of the entity detector is $\lambda_1$ = 2, the weight for predicate prediction is $\lambda_2$ = 1, and the weight for distillation loss is $\lambda_3$ = 20.
All experiments are implemented in PyTorch and trained on 4 NVIDIA A40 GPUs. 

\begin{table*}[ht]
\centering
\setlength{\tabcolsep}{1mm}
\fontsize{9pt}{9pt}\selectfont
\begin{tabular}{l|l|l|l|l|cc|cc|cc}
\toprule
\multirow{2}{*}{S} & \multirow{2}{*}{T}      & \multirow{2}{*}{B} & \multirow{2}{*}{D} & \multirow{2}{*}{M} & \multicolumn{2}{c|}{Total} & \multicolumn{2}{c|}{Novel (Object)} & \multicolumn{2}{c}{Novel (Relation)} \\
                         &                            &                           &                           &                        & R@50         & R@100       & R@50             & R@100            & R@50             & R@100             \\ \midrule
\multirow{4}{*}{SGDet}   & \multirow{2}{*}{t} & \multirow{2}{*}{Swin-T}   & \multirow{2}{*}{-}        & VS$^3$                    & 5.88         & 7.20        & 6.00             & 7.51             & 0.00                & 0.00                 \\
                         &                            &                           &                           & \textbf{VS$^3$+RAHP}                  & \textbf{12.66}        & \textbf{15.39}       & \textbf{13.01}            & \textbf{14.82}            & \textbf{3.75}             & \textbf{5.12}              \\ \cmidrule{2-11} 
                         & \multirow{2}{*}{o} & \multirow{2}{*}{Swin-T}   & \multirow{2}{*}{DETR}     & OvSGTR                 & 13.53        & 16.36       & \textbf{14.37}            & \textbf{17.44}            & 9.20             & 11.19             \\
                         &                           &                           &                           & \textbf{OvSGTR+RAHP}                  &  \textbf{13.83}            &  \textbf{16.52}           &  12.45                &  15.38                &  \textbf{13.31}                &  \textbf{16.46}                 \\ \bottomrule
\end{tabular}%
\caption{Experimental results of OVD+R SGG on VG test set. S is the SGG setting; T denotes the SGG model type, o means one-stage model, t is the two-stage model; B is the backbone model; D is the object detector; M represents the model.}
\label{tab: OVDR SGG-table}
\end{table*}

\begin{table*}[ht]
\centering
\setlength{\tabcolsep}{1mm}
\fontsize{9pt}{9pt}\selectfont
\begin{tabular}{l|lll|cc|cc|cc}
\toprule
\multirow{2}{*}{\#} & \multirow{2}{*}{EP} & \multirow{2}{*}{RP} & \multirow{2}{*}{DS} & \multicolumn{2}{c|}{Total (Relation)} & \multicolumn{2}{c|}{Base (Relation)} & \multicolumn{2}{c}{Novel (Realtion)} \\
                    &                                      &                                      &                                    & R@50/100          & mR@50/100         & R@50/100          & mR@50/100        & R@50/100          & mR@50/100        \\ \midrule
1                   &                                      &                                      &                                    & 21.66/25.89       & 8.49/10.84        & 21.85/26.06       & 9.11/12.42       & 6.15/9.88         & 4.73/8.13        \\
2                   & $\surd$                                    &                                      &                                    & 23.78/28.88       & 9.62/13.65        & 23.97/29.06       & 10.71/14.65      & 8.62/13.05        & 7.09/11.10       \\
3                   &                                      & $\surd$                                    &                                    & 23.81/28.04       & 9.01/11.99        & 23.49/28.91       & 10.10/13.08      & 2.38/5.44         & 1.33/3.77        \\
4                   &                                      & $\surd$                                    & $\surd$                                  & 23.71/27.98       & 9.43/12.51        & 23.95/28.24       & 10.55/14.38      & 6.20/9.10         & 6.94/9.59        \\
5                   & $\surd$                                    & $\surd$                                    &                                    & 22.20/26.07       & 9.33/13.11        & 24.13/28.85       & 10.80/15.08      & 4.25/6.44         & 3.88/5.25        \\
6                   & $\surd$                                    & $\surd$                                    & $\surd$                                  & \textbf{24.25/29.17}       & \textbf{10.09/13.85}       & \textbf{24.43/29.35}       & \textbf{11.46/15.39}      & \textbf{9.25/13.44}        & \textbf{8.88/11.35}       \\ \bottomrule
\end{tabular}%
\caption{Ablation study on model components of VG val set. EP: Entity-aware Prompt; RP: Region-aware Prompt; DS: Dynamic Selection. The first row represents the baseline with a fixed predicate text prompt.}
\label{tab:ablation-study}
\end{table*}

\subsection{Comparisons with OVR-SGG Methods}
\subsubsection{Setup}
We evaluate our design on the VG and OIv6 datasets, comparing it with OVR-SGG methods, including SVRP, Epic, PSGS, and OvSGTR (see Table~\ref{tab:main-table}). We adapt SGG methods (SGTR$^\dagger$ and PE-NET) for OVR-SGG, as they perform well in closed-vocabulary settings. In each method, we retain the visual module and replace the predicate classifier with RAHP's OV predicate prediction module. Additionally, we use a fixed text prompt baseline for comparison, where the prompt only provides information about predicate categories. In the VG dataset's PredCLS setting, we follow Epic's predicate split, selecting 70\% of the categories as base predicates and the remaining 30\% as novel predicates. In the SGDet setting, we follow the OvSGTR predicate split. For the OIV6 dataset, we use the predicate split from PGSG. During training, only base relation annotations are available, with images lacking base relation annotations masked.

\subsubsection{Visual Genome}
Compared to previous methods in Table~\ref{tab:main-table}, our approach demonstrates significant performance advantages. For instance, in the PredCLS task, our method improves the novel mR@100 by 7.76 over the one-stage method PGSG. In two-stage methods, compared to Epic, our method increases the novel R@100 by 1.75 in the PredCLS task. Whether in one-stage or two-stage methods, RAHP shows flexible generalizability, is capable of achieving open-vocabulary capabilities under different frameworks.
Taking PE-NET as an example, our method outperforms baseline models with fixed text prompts, improving both base and novel predicate performance. This highlights RAHP's ability to enhance text representations and improve visual relation understanding.

\subsubsection{Open Image v6}
We compare our method with PGSG on the OIV6 dataset. As shown in Table~\ref{tab:OIV6-table}, our approach achieves an improvement of 21.56 points in total R@100 and 25.26 points in novel mR@100. This demonstrates that the introduction of hierarchical text prompts can enhance text representation, leading to better visual-text matching. 

\subsection{Comparisons with OVD+R-SGG Methods}

\subsubsection{Setup}
We evaluate the performance of RAHP in a fully open vocabulary setting OVD+R-SGG on VG, where novel object and relationship categories are excluded during the training phase. Additionally, we achieve fully open vocabulary capability on $\mathrm{VS^3}$ by replacing the original closed-set predicate classifier with RAHP. We the performance of OvD+R SGG, covering results in three aspects: ``Total'' (i.e., all object and relationship categories), ``Novel (Object)'' (i.e., considering only novel object categories), and ``Novel (Predicate)'' (i.e., considering only novel predicate categories)."

\subsubsection{Results} Table~\ref{tab: OVDR SGG-table} shows that the inclusion of RAHP significantly improved the performance of novel relation, whether in $\mathrm{VS^3}$ or OvSGTR. RAHP expands the text representation space by dynamically selecting region-aware text prompts. This enhances the model's generalization ability, making it more effective in handling new relationship concepts.
Compared the OvSGTR in line 3, RAHP increases R@100 by 5.27. 
We observe a performance drop on novel objects, likely due to the differing distillation methods of the two models: OvSGTR uses relation feature distillation from a pre-trained model, while RAHP employs visual feature distillation from VLM. These differences may lead to conflicts between the approaches.

\subsection{Ablation Study}
We conduct an ablation study to assess the impact of each part on the method's effectiveness and the validity of SGG training. We divide RAHP into three main components: the entity-aware prompt, the region-aware prompt, and the dynamic selection mechanism. We analyze their roles individually based on PE-NET under the OVR-SGG SGDet setting. The results are summarized in Table~\ref{tab:ablation-study}.

Replacing the fixed predicate text prompt with the entity-aware prompt results in a 3\% performance improvement for both base and novel predicates, highlighting the effectiveness of incorporating entity information in enhancing text representation of relations. This underscores the importance of relation triplet information in relationship detection. Introducing the region-aware prompt further enhances text representation, improving alignment between visual and textual features. However, without a dynamic selection mechanism to filter region-aware prompts, performance on novel predicates declines due to noise interference from irrelevant prompts. Base predicates, with inherently higher prediction scores, are more robust against this noise. Implementing an image-guided filtering strategy effectively removes noise, improving prediction accuracy for novel predicates. 

\section{Conclusion}
\label{sec:conclusion}


In this paper, we introduce the Relation-Aware Hierarchical Prompting framework (RAHP), designed to address the challenges of OV-SGG by enhancing text representations. By integrating entity-aware and region-aware relation text prompts, RAHP enhances text representation and enables more accurate and flexible image-text matching. Our dynamic selection mechanism further refines this process by adapting prompts based on visual information, reducing noise and improving the robustness of relation predictions.
Through extensive experiments on the Visual Genome and Open Images v6 datasets, our method demonstrates state-of-the-art performance, and the demonstrated performance improvements—highlight the potential of RAHP to significantly advance the field of OV-SGG.

\textbf{Discussion of Limitations}:
(1) \textit{Effectiveness of Entity Clustering}. 
Clustering algorithms will struggle to maintain fine distinctions between diverse data categories, which can degrade the quality of text representations.
(2) \textit{Diversity of Generated Text Prompts}. Limited diversity in LLM-generated region descriptions can hinder model generalization for novel relationships.

\section{Acknowledgments}
This work was supported by NSFC 62350610269, Shanghai Frontiers Science Center of Human-centered Artificial Intelligence, and MoE Key Lab of Intelligent Perception and Human-Machine Collaboration (ShanghaiTech University).



\newpage
\appendix
\section{Overview of Material}
In this supplementary document, we provide additional details and experimental results to enhance understanding and insights into our proposed RAHP. This supplementary document is organized as follows:
\begin{itemize}
    \item  A detailed description of the entity clustering method in Sec.~\ref{sup:sec-1} is provided.

    \item Examples generated by the region-aware prompt mining in Sec.~\ref{sup:sec-2} are presented.

    \item Detailed descriptions of the Visual Genome and Open Image V6 datasets used in our study are provided in Sec.~\ref{sup:sec-3}.

    \item Additional experimental results are presented in Sec.~\ref{sup:sec-4}.
\end{itemize}

\section{Entity Clustering}
\label{sup:sec-1}
\subsection{Entity Clustering Method}
In RAHP, we employ entity clustering to reduce the number of relationship triplets, thereby preventing the inefficiency in text prompt usage caused by the proliferation of triplets. 
To define superclasses, we leverage the lexical structure and part-of-speech tagging provided by WordNet. Specifically, we first cluster entities using WordNet’s part-of-speech tags, grouping semantically related terms together. Next, we encode each cluster using the text encoder of a Vision-Language Model (VLM) and apply the K-means algorithm to the encoded embeddings, setting the number of clusters to M, resulting in M distinct categories.

To generate names for each super entity and minimize ambiguity, we utilize the extensive knowledge of a Large Language Model (LLM) to select the most frequent or semantically relevant superclasses. This approach enhances the stability and representativeness of the superclasses. We input all entity categories within each cluster into the LLM sequentially, prompting it to generate appropriate super entity names. To preserve the key characteristics of the entities, we instruct the LLM to analyze essential attributes before generating the super entity name.
The prompt for generating the superclasses is shown is as follows:

\textcolor{orange}{
Task Description: You will be provided with a set of predicates related to specific actions, states, or relationships. Your task is to generate an appropriate superclass category name that effectively encapsulates the common characteristics of these predicates.}
\\[0.5cm]
\textcolor{orange}{Input: You will receive the following set of predicates.}
\\[0.5cm]
\textcolor{orange}{Output: 
Please provide a concise and specific superclass category name that encompasses all the given predicates. The superclass name should be between one to three words and should use general and easily understandable vocabulary.}

\subsection{Super Entity Categories}
The VG super entities generated by this method are as follows:

VG\_super\_entities = \lstinline|[''male'', ''female'', ''children'', ''pets'', ''wild animal'', ''ground transport'', ''water transport'', ''air transport'', ''sports equipment'', ''seating furniture'', ''decorative item'', ''table'', ''upper body clothing'', ''lower body clothing'', ''footwear'', ''accessory'', ''fruit'', ''vegetable'', ''prepared food'', ''beverage'', ''utensils'', ''container'', ''textile'', ''landscape'', ''urban feature'', ''plant'', ''structure'', ''household item'', ''head part'', ''limb and appendage'']|

The resulting OIV6 super entities are as follows:

OIV6\_super\_entities = \lstinline|[''male'', ''female'', ''children'', ''head feature'', ''limb feature'', ''torso feature'', ''accessorie'', ''mammal'', ''bird'', ''reptile'', ''insect'', ''marine animal'', ''bike'', ''ground vehicle'', ''watercraft'', ''aircraft'', ''vehicle part item'', ''ball-related sport item'', ''water sport item'', ''winter sport item'', ''seating furniture'', ''table furniture'', ''storage furniture'', ''bedding'', ''upper body clothing'', ''lower body clothing'', ''footwear'', ''fruit'', ''vegetable'', ''prepared food'', ''beverage'', ''appliance'', ''utensil'', ''decorative item'', ''textile'', ''hand tool'', ''power tool'', ''kitchen tool'', ''personal electronic'', ''home electronic'', ''office electronic'', ''land vehicle'', ''water vehicle'', ''air vehicle'', ''string instrument'', ''wind instrument'', ''percussion instrument'', ''firearm'', ''container'', ''toy'', ''stationery'', ''landscape'', ''urban feature'']|

\section{Region-aware Prompts Generation}
\label{sup:sec-2}
\subsection{Region-aware Prompts Statistics}
This section gives an example of generating a region description. We use LLM extract key parts from both the subject and object, combining these parts to create detailed region descriptions. This approach allows the LLM to pinpoint the exact regions where subject-object interactions occur, leading to more precise visual relation descriptions for each relationship triplet. 
Using this prompt, we generate region descriptions for relationship triplets, achieving an average of 6.76 prompts per triplet with 20.32 unique objects and 7.58 unique relations, significantly outperforming baselines with only 2 objects and 1 relation. Empirically, our approach improves performance across 97 object classes in VG, highlighting its generalizability beyond specific categories. Here is a region description for the relationship triplet $<$vegetable, in, container$>$.

\subsection{Region-aware Prompts Generation Examples}
The complete region descriptions prompt is as follows:
\\[0.5cm]
\textcolor{orange}{
Describe [subject] [predicate] [object] which parts of subject and object function in this relationship. Please list these parts, and then analyze and describe the visual relationship between these parts.
The generated description should be concise and clear. Here are two examples for you to learn:}
\\[0.5cm]
\textcolor{orange}{\hspace{0.35cm}Example A: ``[human] [holding] [wild animal]'': }

\textcolor{orange}{Subject Part : [hand, arm, legs, ...]}

\textcolor{orange}{Object Part : [animal limbs, animal body, ...]}

\textcolor{orange}{Region Rescriptions :}

\textcolor{orange}{[``human hand(s) securely gripping the animal'', ``human arm(s) embracing or supporting the animal'', ``animal positioned close to or physically touching the human's torso'', ``animal appears stable and not struggling'', ``direct gaze or interaction between the human and the animal suggesting control or care'', ``human fingers intertwined or wrapped around the animal's body or limbs'', ``animal's posture conveys being held, often with limbs tucked or supported'',   ``proximity of the human face to the animal, especially when holding smaller animals'', ``human holding the animal with hands'', ``human's hands or arms in contact with the animal'', ``animal is held in the human's arms'']}
\\[0.5cm]
\textcolor{orange}{\hspace{0.35cm}Example B: ``[human] [sitting on] [seating furniture]'':}

\textcolor{orange}{Subject Part : [buttocks, thighs, legs, back, arms]}

\textcolor{orange}{Object Part : [seat, backrest, armrests]}

\textcolor{orange}{Region Rescriptions :}

\textcolor{orange}{[``Human's buttocks are making contact with the seat of the furniture.'', ``Human's thighs rest on the seat, with legs positioned either bent or extended.'', ``Human's back is supported by the backrest of the furniture.'', ``Human's arms may be resting on or near the armrests of the furniture, if present.'', ``The furniture's seat aligns with the human's buttocks and thighs, indicating proper seating support.'', ``The human's posture is influenced by the backrest, which can be either upright or reclining.'', ``The armrests, if present, support the human's arms, enhancing comfort and stability.'', ``The arrangement of the human's legs and feet suggests their interaction with the seat and alignment with the furniture.'']}


\definecolor{lightblue}{rgb}{0.23, 0.65, 0.75}
\textcolor{lightblue}{\hspace{0.35cm}``[vegetable] [in] [container]'': }

\textcolor{lightblue}{Subject Part : [vegetable, leaves, stems, roots]}

\textcolor{lightblue}{Object Part : [container, soil, water]}

\textcolor{lightblue}{Region Rescriptions :}

\textcolor{lightblue}{["The vegetable is contained within the boundaries of the container, with its roots potentially reaching into the soil or water for nutrients." "The leaves and stems of the vegetable are positioned within the confines of the container, adapting to the available space for growth." "The container provides a stable environment for the vegetable, supporting its growth and development," "The vegetable's roots may be intertwined or spread out within the container, seeking out moisture and nutrients." "The visual relationship between the vegetable and the container showcases a symbiotic connection, where the container nurtures and protects the vegetable."] this relationship, the subject part (vegetable) is dependent on the object part (container) for its growth and sustenance. The visual relationship between these parts highlights the interaction and support provided by the container to the vegetable, creating a harmonious and nurturing environment for the vegetable to thrive. The container acts as a protective barrier, while also serving as a source of nutrients and stability for the vegetable. The roots of the vegetable reaching into the soil or water within the container further emphasize the interconnectedness and reliance of the vegetable on its container for survival."]}

\section{Datasets and Relation Split}
\label{sup:sec-3}
We evaluate our method on the Visual Genome (VG) and Open Image v6 (OIV6) datasets. This section details each dataset and our approach to dividing base and novel relations during experiments.

\textbf{Visual Genome}: VG includes 150 entity categories and 50 predicate categories, manually annotated across 108,777 images. We use 70\% of the images for training, 5,000 for validation, and the remainder for testing.

PredCLS Setting: Following Epic’s method, we categorize 70\% of the predicates as base predicates and the remaining 30\% as novel predicates. The base predicates are: \lstinline|[''above'', ''against'', ''at'', ''attached to'', ''behind'', ''belonging to'', ''between'', ''carrying'', ''covered in'', ''covering'', ''for'', ''from'', ''hanging from'', ''has'', ''holding'', ''in'', ''in front of'', ''looking at'', ''made of'', ''near'', ''of'', ''on'', ''over'', ''parked on'', ''playing'', ''riding'', ''sitting on'', ''standing on'', ''to'', ''under'', ''walking on'', ''watching'', ''wearing'', ''wears'', ''with'']|

The novel predicates are: \lstinline|[''across'', ''along'', ''and'', ''eating'', ''flying in'', ''growing on'', ''laying on'', ''lying on'', ''mounted on'', ''on back of'', ''painted on'', ''part of'', ''says'', ''using'', ''walking in'']|

SGDet Setting: We adopt OvSGTR's division of predicates, 
the divided base predicates are: \lstinline|["between", "to", "made of", "looking at", "along", "laying on", "using", "carrying", "against", "mounted on", "sitting on", "flying in", "covering", "from", "over", "near", "hanging from", "across", "at", "above", "watching", "covered in", "wearing", "holding", "and", "standing on", "lying on", "growing on", "under", "on back of", "with", "has", "in front of", "behind", "parked on"]|
and the novel predicates are: \lstinline|["belonging to", "part of", "riding", "walking in", "in", "of", "painted on", "playing", "for", "walking on", "says", "attached to", "eating", "on", "wears"]|

\textbf{Open Image v6}: OIV6 contains 301 entity categories and 31 predicate categories. We use 126,368 images for training, 1,813 for validation, and 5,322 for testing.
Following PGSG's method, we split all predicates to base and novel predicates two parts.

Base predicates: \lstinline|['at', 'holds', 'wears', 'holding hands', 'on', 'highfive', 'contain', 'handshake', 'talk on phone']|

Novel predicates: \lstinline|['surf', 'hang', 'drink', 'ride', 'dance', 'skateboard', 'catch', 'inside of', 'eat', 'cut', 'kiss', 'interacts with', 'under', 'hug', 'throw', 'hits', 'snowboard', 'kick', 'ski', 'plays', 'read']|

\section{More Experimental Results}
\label{sup:sec-4}
\subsection{Close-vocabulary SGG on VG}
\label{exp: result-1}
In Tab.~\ref{tab:close-set-table}, we present the close-vocabulary SGDet performance on VG.
We compare the performance of five models mentioned in the experimental section: PGSG, PE-NET, VS$^3$, SGTR$^\dagger$. The results indicate that substituting the fixed classifier with RAHP's open-vocabulary classifier enhances both recall and mean recall in PE-NET and SGTR$^\dagger$. For VS$^3$, the performance remains comparable. This demonstrates RAHP's robustness in the both close-vocabulary and open-vocabulary SGG task. Moreover, as an open-vocabulary relation classifier, RAHP can be seamlessly extended to open-vocabulary scenarios by expanding its vocabulary, even after training on closed-vocabulary tasks.

\begin{table}[]
\centering
\resizebox{\linewidth}{!}{%
\begin{tabular}{l|l|l|l|ll}
\toprule
T                  & B                       & D                            & M           & R@50/100             & mR@50/100   \\ \midrule
\multirow{4}{*}{t} & \multirow{2}{*}{R-101}  & \multirow{2}{*}{\shortstack{Faster \\ R-CNN}} & PE-NET      & 30.70/35.20          & 12.40/14.50 \\
                   &                         &                              & PE-NET+RAHP & 31.52/36.46          & 12.75/15.21 \\ \cmidrule{2-4}
                   & \multirow{2}{*}{Swin-T} & \multirow{2}{*}{-}           & VS$^3$         & \textbf{35.80/41.30} & -           \\
                   &                         &                              & VS$^3$+RAHP    & 34.25/40.40          & 7.21/10.45  \\ \midrule
\multirow{3}{*}{o} & ViT                     & DETR*                        & PGSG        & 16.70/21.20          & 8.90/11.50  \\ \cmidrule{2-4}
                   & \multirow{2}{*}{R-101}  & \multirow{2}{*}{DETR}        & SGTR$^\dagger$       & 25.80/29.60          & 12.60/17.00 \\
                   &                         &                              & SGTR$^\dagger$+RAHP  & \textbf{25.95/29.60} & 12.49/17.04 \\ \bottomrule
\end{tabular}%
}
\caption{The close-vocabulary SGDet performance on VG, DETR* denotes models with structural modifications. t means the two-stage model, o means the one-stage model.}
\label{tab:close-set-table}
\end{table}

\subsection{More Experimental Results of SGCLS}
In addition to the PredCLS and SGDet protocols, we conduct OVR-SGG experiments using SGTR+ as the backbone model under the SGCLS protocol. The results are summarized in the Table~\ref{tab:3}. Compared to SGTR+, RAHP enhances the model’s open-vocabulary capabilities, effectively recognizing novel predicates. Additionally, RAHP offers a rich set of prompts, enabling better differentiation among various predicates and improving the model’s prediction accuracy for novel predicates.

\begin{table*}[]
\centering
\caption{Experimental results of OVR-SGG in the VG test set and SGCls protocol, p indicates the use of fixed-format text prompts.}
\label{tab:3}
\resizebox{0.8\textwidth}{!}{%
\begin{tabular}{c|c|cc|cc|cc}
\toprule
\multirow{2}{*}{S}     & \multirow{2}{*}{M} & \multicolumn{2}{c|}{Total (Relation)}       & \multicolumn{2}{c|}{Base (Relation)}        & \multicolumn{2}{c}{Novel (Relation)}        \\
                       &                    & R@50/100             & mR@50/100            & R@50/100             & mR@50/100            & R@50/100             & mR@50/100           \\ \midrule
\multirow{3}{*}{SGCLS} & SGTR$^\dagger$              & 21.42/26.60          & 9.13/11.58           & 32.18/36.11          & 13.12/16.88          & 0.00/0.00            & 0.00/0.00           \\
                       & SGTR$^\dagger$+p            & 25.98/29.31          & 11.28/15.39          & 32.51/36.55          & 13.24/17.01          & 11.03/18.52          & 6.70/11.61          \\
                       & SGTR$^\dagger$+RAHP         & \textbf{26.65/30.83} & \textbf{12.98/16.81} & \textbf{32.63/36.83} & \textbf{13.86/17.14} & \textbf{14.37/21.72} & \textbf{9.91/15.04} \\ \bottomrule
\end{tabular}%
}
\end{table*}

\subsection{Impact of Hyper-parameter $\alpha$}
\label{exp: result-2}
\begin{figure}[t]
  \centering
  \includegraphics[width=\linewidth]{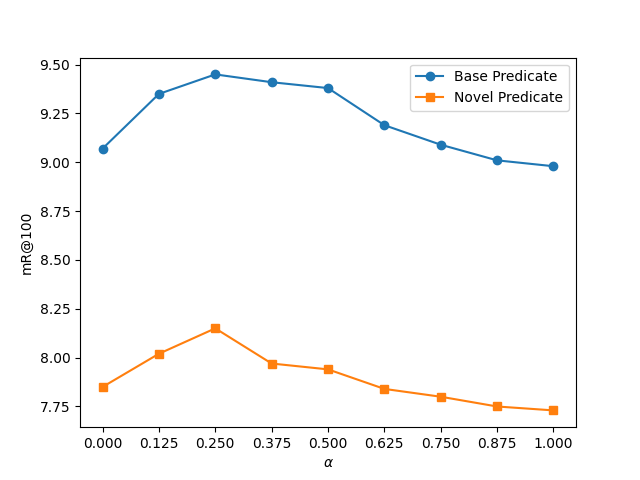}
  \caption{Impact of Hyper-Parameter $\alpha$ on RAHP Performance in the VG validation set.
  }
  \label{exp:alpha_fig}
\end{figure}

We conduct experiments on the VG validation set to evaluate the impact of different parameters $\alpha$, which controls the weight of the region-aware prompt in predicate score aggregation. The SGG model used in this experiment is PE-NET. 

As shown in Fig.~\ref{exp:alpha_fig}, the results indicate that the model performs best when $\alpha=0.25$. In contrast, both $\alpha=0$ (no region-aware prompt) and $\alpha=1$ (no entity-aware prompt) lead to a decline in performance for both base and novel predicates. 

\subsection{Impact of Hyper-parameter $k$}
\label{exp: result-3}
\begin{figure}[t]
  \centering
  \includegraphics[width=\linewidth]{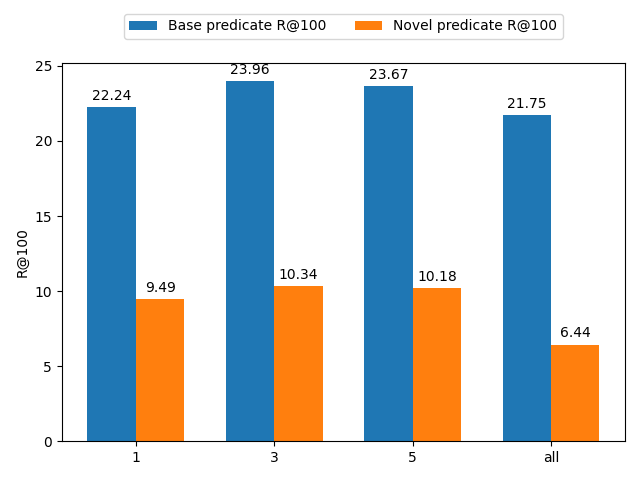}
  \caption{Impact of Hyper-Parameter $k$ on RAHP Performance in the VG validation set.
  }
  \label{exp:k_fig}
\end{figure}

Table.~\ref{exp:k_fig} illustrates the impact of varying the hyperparameter $k$ on model performance, where $k$ represents the number of region-aware prompts selected in the dynamic selection mechanism. Similar to sec.~\ref{exp: result-2} setting, we show the R@100 results on the VG validation set. The model achieves optimal performance at $k=3$. Consistent with our ablation study, selecting all region-aware prompts introduces noise, degrading performance for novel predicates. Conversely, limiting the selection to only the highest-scoring (Top 1) prompts reduces the diversity of text representations, thereby impairing SGG performance.

\subsection{Impact of Distillation loss weight $\lambda_3$}
\label{exp: result-4}
\begin{figure}[t]
  \centering
  \includegraphics[width=\linewidth]{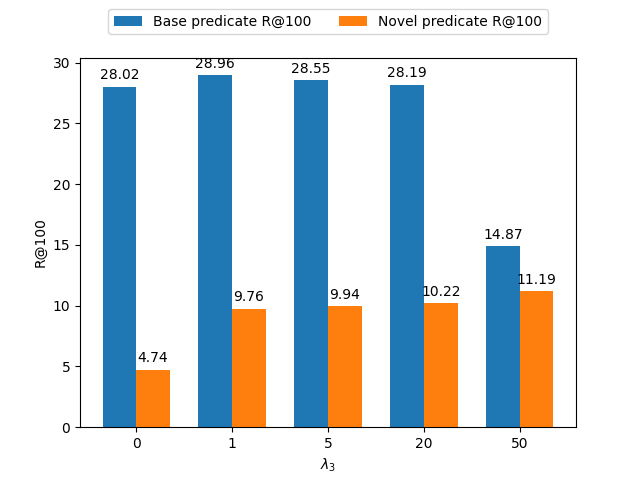}
  \caption{Impact of Distillation Loss Weight $\lambda_3$ on RAHP Performance in the VG validation set.
  }
  \label{exp:distillation_fig}
\end{figure}

\begin{figure}[t]
  \centering
  \includegraphics[width=\linewidth]{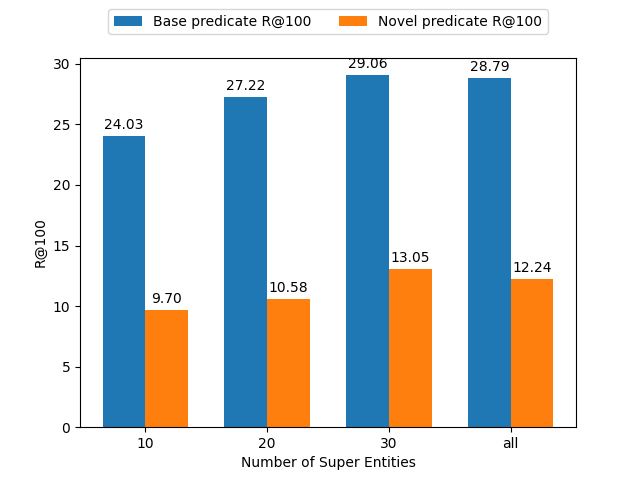}
  \caption{Impact of the super entity's number on RAHP performance in the VG validation set.
  }
  \label{exp:super_fig}
\end{figure}

We discuss the impact of the distillation loss weight $\lambda_3$, on the model's performance, particularly its effect on the alignment between the SGG model's relation representations and the VLM's visual representations. We test four different values of $\lambda_3=(1, 5, 20, 50)$ to determine the optimal setting.

Table~\ref{exp:distillation_fig} presents the results for different $\lambda_3$ values. The findings indicate that the model performs best with $\lambda_3=20$, where the distillation method successfully balances the knowledge learned from SGG data and the knowledge transferred from VLM. This balance allows the model to maintain high performance across both base and novel predicates. 
Without distillation, the model struggles, particularly with novel categories, highlighting the importance of this strategy in retaining essential knowledge while generalizing to novel relationships.
The results at $\lambda_3=50$ demonstrate that while the model can be tuned to improve the recognition of novel relationships, there is a threshold beyond which further increases in $\lambda_3$ may begin to degrade the model's overall effectiveness by disproportionately affecting base predicate performance.

\subsection{Impact of the Number of Super Entities}
\label{exp-ref: result-4}
We explore the impact of varying the number of super entities. Specifically, in the VG dataset, we group entities into super entities of 10, 20, and 30 categories, reflecting different levels of granularity. Fewer super entities indicate more aggregation, leading to less retention of the original entities' features. For example, classifying "bike" under "vehicle" may result in the loss of information specific to "bike" in region descriptions. Thus, the choice of the number of super entities balances between information completeness and computational complexity. Additionally, we conduct experiments on triplets formed by all entities (resulting in a total of 1,125,000 relation triplets, shown as "all" in the Fig.~\ref{exp:super_fig}) to compare the performance when complete entity information is provided.

The Fig.~\ref{exp:super_fig} shows that the number of super entities has a significant impact on model performance. When the number of super entities is limited to 10, the categories become overly broad, leading to a substantial loss of unique information when entities are converted into super entities. This makes it difficult to align the text representations with the visual representations. 
Increasing the number to 30 enhances the granularity of the text representations, improving the matching performance. Interestingly, at this level, the model's performance nearly matches that achieved using full triplets. This finding suggests that representative feature descriptions, even without highly detailed visual features, can substantially boost text-visual matching accuracy.

\begin{table*}[t]
\centering
\caption{Impact of the LLM on RAHP performance in the VG validation set.}
\label{tab:4}
\resizebox{0.8\textwidth}{!}{%
\begin{tabular}{c|c|c|cc|cc|cc}
\toprule
\multirow{2}{*}{S}       & \multirow{2}{*}{LLM} & \multirow{2}{*}{M} & \multicolumn{2}{c|}{Total (Relation)}       & \multicolumn{2}{c|}{Base (Relation)}        & \multicolumn{2}{c}{Novel (Relation)}         \\
                         &                      &                    & R@50/100             & mR@50/100            & R@50/100             & mR@50/100            & R@50/100             & mR@50/100            \\ \midrule
\multirow{3}{*}{PredCLS} & -                    & PE-NET             & 58.79/61.23          & 19.18/20.97          & 63.62/67.09          & 23.18/25.79          & 0.00/0.00            & 0.00/0.00            \\
                         & GPT-3.5-turbo              & PE-NET+RAHP        & 64.70/69.11          & 24.50/28.25          & 65.15/70.54          & 24.99/30.19          & 20.79/29.00          & 15.70/23.73          \\
                         & CPT-4o-mini          & PE-NET+RAHP        & \textbf{64.73/69.15} & \textbf{24.52/28.25} & \textbf{65.14/70.55} & \textbf{25.01/30.17} & \textbf{20.88/29.06} & \textbf{15.73/23.77} \\ \bottomrule
\end{tabular}%
}
\end{table*}

\subsection{Impact of the Large Language Models}
In RAHP, the region-aware prompts are generated using a Large Language Model (LLM). To ensure that variations in different LLMs do not significantly impact RAHP’s performance, we conducted a comparative experiment using GPT-4o-mini. We chose GPT-4o-mini due to its rapid deployment capabilities through the OpenAI API, which offers advantages for method migration and implementation.
The total cost of generating region-aware prompts for 45,000 relation triplets using both GPT-3.5-turbo and GPT-4o-mini is [insert cost here]. As shown in the Table~\ref{tab:4}, experiments conducted on the VG dataset in an OVR-SGG setting revealed that both GPT-3.5-turbo and GPT-4o-mini achieved comparable performance. This comparative study demonstrates that RAHP exhibits robustness across different LLMs, yielding stable performance despite variations in the underlying model.

\subsection{Comparison with Zero-shot Relation Detection Method}
\label{exp: result-5}
To clarify the differences between RAHP and RECODE, we conducted tests on the zero-shot visual relationship detection task. Consistent with RECODE, we utilized CLIP with the Vision Transformer (ViT-B/32) as the default backbone and employed GPT-3.5-turbo as the LLM. The results on the VG dataset as shown in the Table~\ref{tab2}, RAHP achieves significant improvements of 0.6 points in R@20 and 1.7 points in mR@20. The increase in mR indicates that our regional-level descriptions provide richer representations, leading to better performance.

\begin{table}[t]
\centering
\caption{Performance comparison of RAHP and RECODE on Predicate Classification metrics within the zero-shot visual relationship detection task.}
\label{tab2}
\resizebox{\linewidth}{!}{%
\begin{tabular}{l|l|cc}
\toprule
\multirow{2}{*}{Backbond} & \multirow{2}{*}{Method} & \multicolumn{2}{c}{Predicate Classification}                                                                                                                                                                                   \\
                          &                         & \multicolumn{1}{c|}{R@20/50/100}                                                                                         & mR@20/50/100                                                                                        \\ \midrule
\multirow{2}{*}{CLIP}     & RECODE                  & \multicolumn{1}{c|}{10.60 / 18.3 / 25.0}                                                                                 & 10.70 / 18.70 / 27.8                                                                                \\ \cline{2-4} 
                          & RAHP                    & \multicolumn{1}{c|}{\textbf{11.06} / \textbf{18.67} / \textbf{25.93}} & \textbf{12.48} / \textbf{19.65} / \textbf{28.79} \\ \bottomrule
\end{tabular}%
}
\end{table}

\subsection{Qualitative Results}
\label{ref-exp: result-5}
\begin{figure*}[ht]
  \centering
  \includegraphics[width=\textwidth]{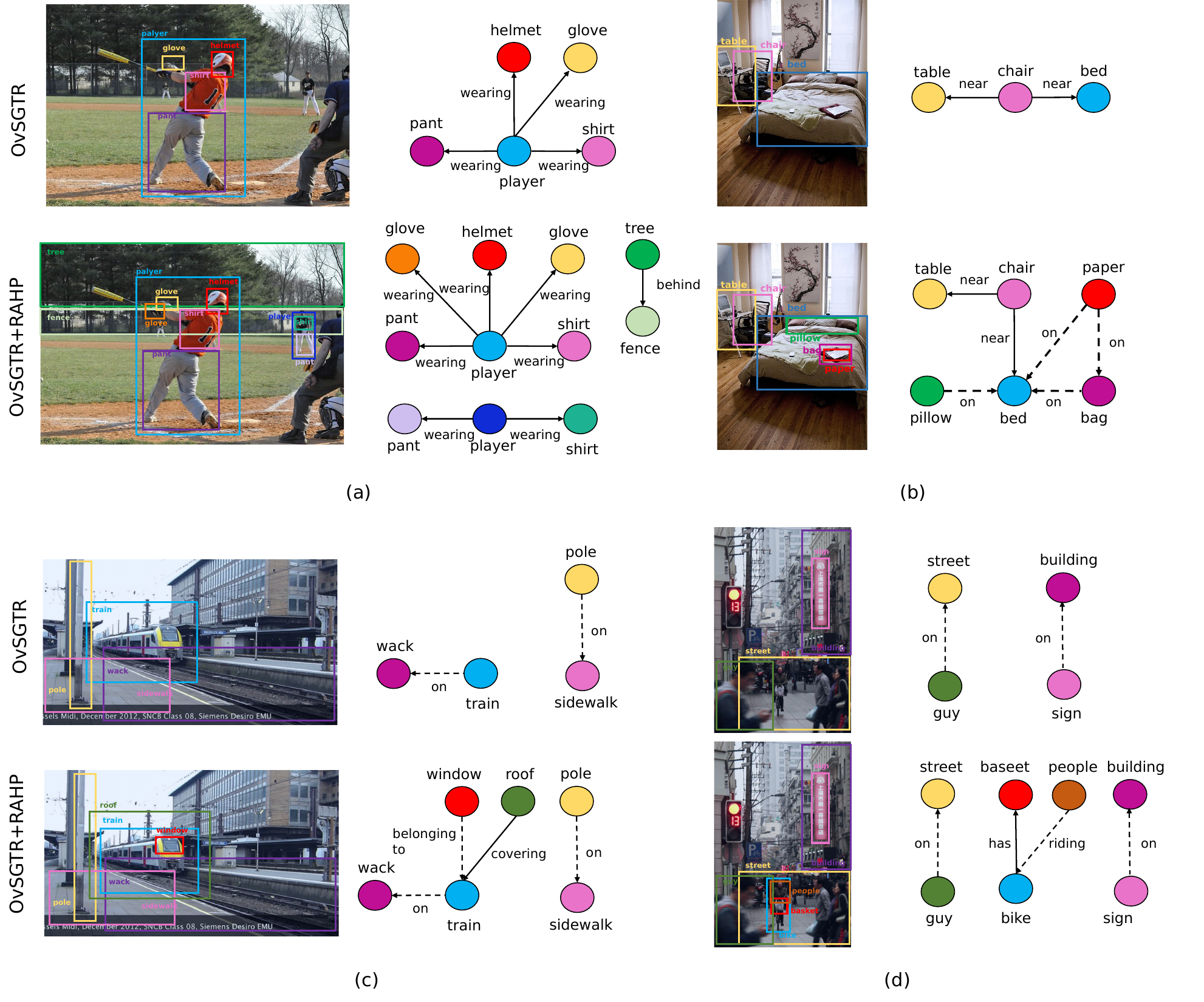}
  \caption{Qualitative  Results of OvSGTR and RAHP on the SGDet task and VG dataset.
  }
  \label{exp:vis_fig}
\end{figure*}

To provide further insights into the effectiveness of our method, we visualize some scene graphs generated by our method and the baseline OvSGTR method in Fig~\ref{exp:vis_fig}. 
The solid lines in the figure represent base predicates, while the dashed lines denote novel predicates.  Compared to the baseline, our method effectively identifies novel relationships and generates richer scene graphs.  RAHP enhances the depth of scene understanding, demonstrating the superiority of our approach in generating scene graphs for complex visual scenarios.

\end{document}